\documentclass[letterpaper,twocolumn]{article}
\usepackage{times} 
\usepackage{helvet} 
\usepackage{courier} 
\usepackage[hyphens]{url} 
\usepackage{graphicx} 
\usepackage{graphicx} 
\usepackage{natbib} 
\usepackage{caption} 
\usepackage[utf8]{inputenc}   
\usepackage[symbol]{footmisc} 

\usepackage[colorlinks=true,   
            linkcolor=black,    
            citecolor=black,    
            urlcolor=black,      
            breaklinks=true]{hyperref}
\usepackage{url}   
\newcommand{\hrefurl}[2]{\href{#1}{\texttt{#2}}}

\usepackage{amsfonts, amsmath, amssymb}
\usepackage{algorithm}
\usepackage{algorithmic}

\usepackage{natbib}
\setcitestyle{round}

\usepackage[most]{tcolorbox} 

\newtcolorbox[auto counter]{floatingexample}[2][]{%
  float,
  floatplacement=t,
  title={Box~\thetcbcounter: #2},
  label={#1},
  colback=gray!0!white,
  colframe=gray!25!white,
  fonttitle=\bfseries,
  coltitle=black,
  width=\columnwidth,
  boxrule=0.8pt,
  arc=0pt,
  outer arc=0pt, 
  left=4pt,
  right=4pt,
  top=2pt,
  bottom=2pt,
}

\newtcolorbox[auto counter]
{floatingexampletwo}[2][]{%
  float,
  floatplacement=h,
  title={Box~2: #2},
  label={#1},
  colback=gray!0!white,
  colframe=gray!25!white,
  fonttitle=\bfseries,
  coltitle=black,
  width=\columnwidth,
  boxrule=0.8pt,
  arc=0pt,
  outer arc=0pt, 
  left=4pt,
  right=4pt,
  top=2pt,
  bottom=2pt,
}
\usepackage{float}
\usepackage{newfloat}
\usepackage{listings}
\usepackage{lipsum}
\DeclareCaptionStyle{ruled}{labelfont=normalfont,labelsep=colon,strut=off} 
\floatstyle{ruled}
\newfloat{listing}{tb}{lst}{}
\floatname{listing}{Listing}

\DeclareMathOperator*{\argmax}{arg\,max}
\DeclareMathOperator*{\argmin}{arg\,min}

\usepackage{cleveref} 
\usepackage{subcaption}
\usepackage{booktabs}
\usepackage{multirow}

\title{Mapping on a Budget: Optimizing Spatial Data Collection for ML}

\author{
  Livia Betti\textsuperscript{1}\footnotemark[1] \and
  Farooq Sanni\textsuperscript{2,3} \and
  Gnouyaro Sogoyou\textsuperscript{2,3} \and
  Togbe Agbagla\textsuperscript{2} \and
  Cullen Molitor\textsuperscript{3,4} \and
  Tamma Carleton\textsuperscript{3,4} \and
  Esther Rolf\textsuperscript{1}
}

\newcommand{\affiliations}{
\textsuperscript{1}University of Colorado Boulder\\
\textsuperscript{2}Togo Data Lab\\
\textsuperscript{3}Center for Effective Global Action\\
\textsuperscript{4}University of California, Berkeley
}

\date{}  

\begin{document}

\twocolumn[
\maketitle
{\centering
\vspace{-1em} 
\affiliations
\par
\vspace{0.5em} 
}
]

\footnotetext[1]{Email: livia.betti@colorado.edu}

\begin{abstract}
In applications across agriculture, ecology, and human development, machine learning with satellite imagery (SatML) is limited by the sparsity of labeled training data. 
While satellite data cover the globe, labeled training datasets for SatML are often small, spatially clustered, and collected for other purposes (e.g., administrative surveys or field measurements). 
Despite the pervasiveness of this issue in practice, past SatML research has largely focused on new model architectures and training algorithms to handle scarce training data, rather than modeling data conditions directly.
This leaves scientists and policymakers who wish to use SatML for large-scale monitoring uncertain about whether and how to collect additional data to maximize performance.
Here, we present the first problem formulation for the optimization of spatial training data in the presence of heterogeneous data collection costs and realistic budget constraints, as well as novel methods for addressing this problem. 
In experiments simulating different problem settings across three continents and four tasks, our strategies reveal substantial gains from sample optimization.
Further experiments delineate settings for which optimized sampling is particularly effective.
The problem formulation and methods we introduce are designed to generalize across application domains for SatML; we put special emphasis on a specific problem setting where our coauthors can immediately use our findings to augment clustered agricultural surveys for SatML monitoring in Togo. Our code is publicly available at \hrefurl{https://github.com/UCBoulder/optimizedsampling}{github.com/UCBoulder/optimizedsampling}.
\end{abstract}


\section{Introduction}

Machine learning with satellite imagery (SatML) is increasingly used to fill crucial gaps in the coverage of existing ground-based data sampling efforts, resulting in global- and national-scale maps of poverty \citep{jean2016poverty}, deforestation \citep{hansen2013high}, cropland \citep{potapov2022global}, and many other outcomes. These maps unlock policy solutions on a wide range of urgent issues, informing design of conservation programs \citep{heilmayr2020brazil}, targeting of aid to those most in need \citep{smythe2022geographic}, and monitoring illegal activity \citep{mcdonald2021satellites}. 

\begin{figure}[t]
\includegraphics[height=4cm]{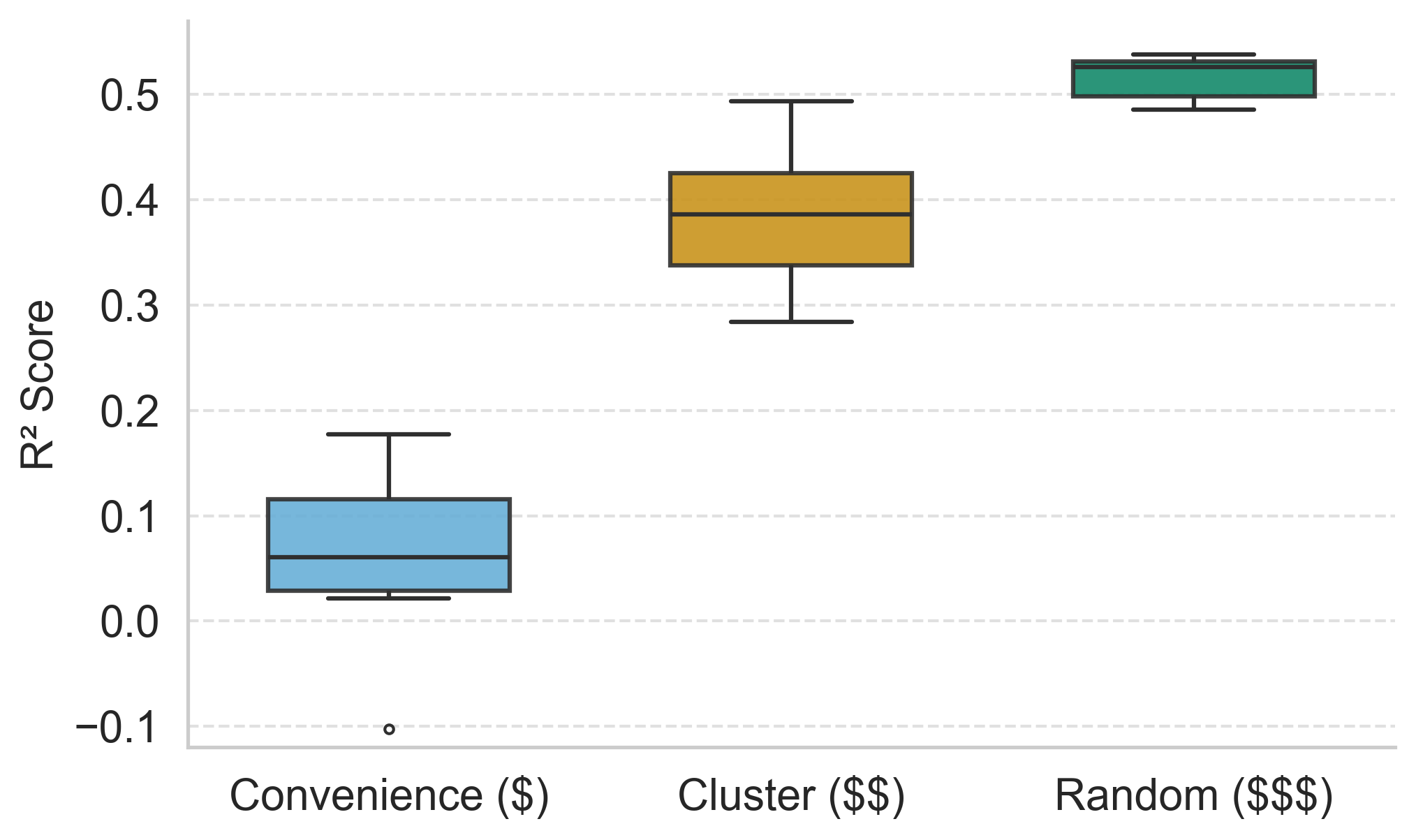} 
\caption{\textbf{Quality and cost of a training dataset depend on its spatial distribution.}
This illustrative figure shows the performance ($R^2$) of a model trained on datasets of sample size of 300 collected according to different sampling methods, for the task of predicting population density in the US. 
}
\label{fig: data collection initial set performance}
\end{figure}

To date, this progress has been fueled by significant methodological advances in ML model architectures and satellite sensor technologies.
In contrast, very little research attention has been paid to methods for training data collection, despite growing consensus that the quantity and spatial distribution of ground-referenced training labels is a consistent barrier to performance of these models in practice \citep{burke2021using,rolf2024mission}. For example, spatially clustered training data limit the generalizability of SatML maps over regions with large data gaps \citep{christensen2025estimating}. As governments and firms increasingly deploy SatML to monitor outcomes critical to human and environmental well-being, no evidence-based guidance exists on how to design sampling methodologies for the datasets used to train these models. Even more importantly, no prior work evaluates how the structure of those datasets interacts with the binding budget constraints that predominate in the data-poor environments where SatML is most impactful. Thus, while the quality and cost of training datasets depend critically on their spatial distribution (\Cref{fig: data collection initial set performance}), policymakers and researchers have no guidance on how to optimally conduct data collection for SatML applications.

To address this gap, \textbf{we formalize a novel problem statement for optimizing spatial collection of training data}, which captures two key features of SatML settings: 

\begin{floatingexample}[float, floatplacement=h, ex:togobox]{Motivating problem setting in Togo}
\begin{center}
\includegraphics[width=1\textwidth]{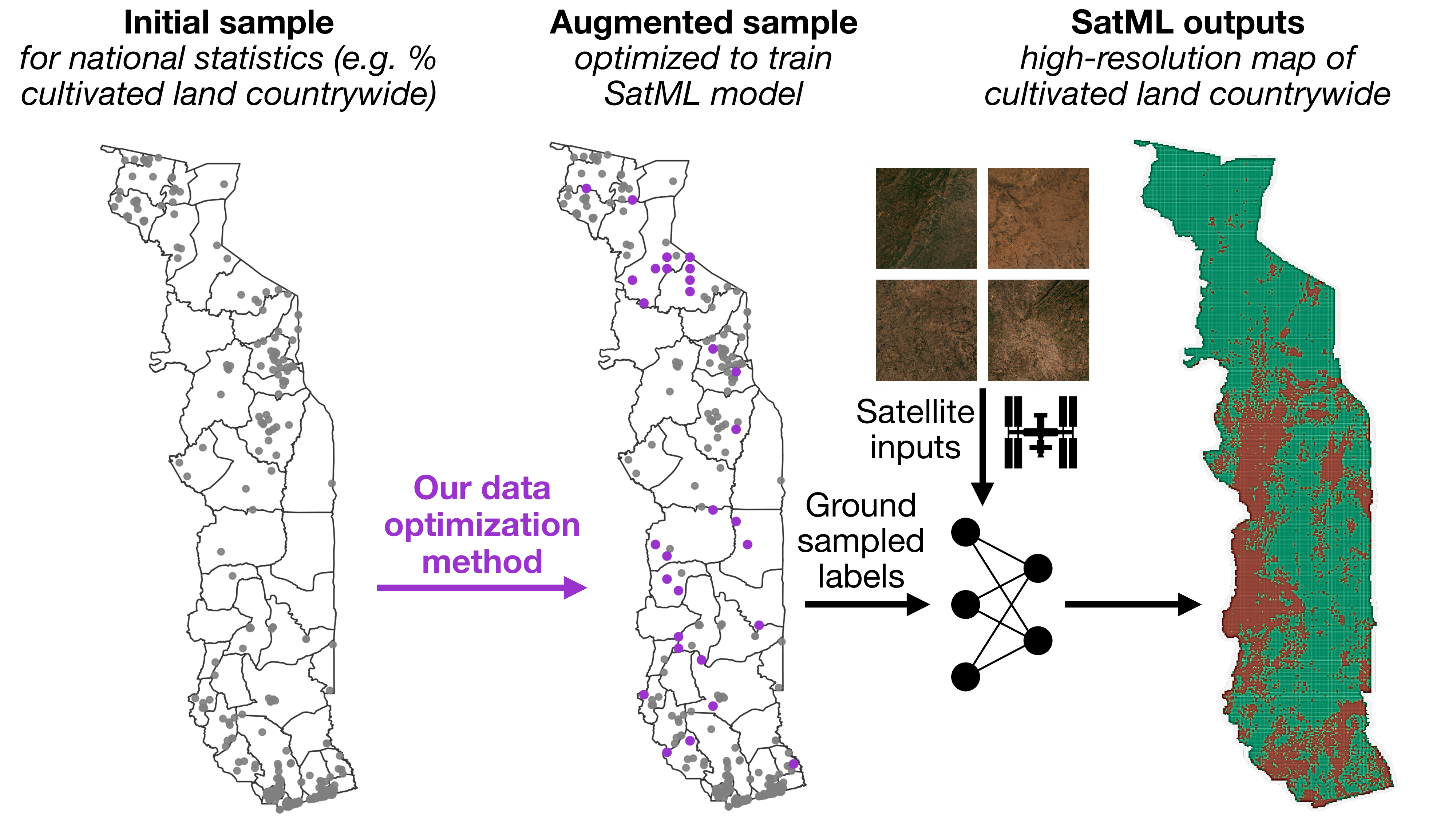}
\captionof{figure}{Optimizing survey augmentation for SatML.}
\label{fig:data_setup_togo}
\end{center}
Togo's Ministry of Agriculture (MinAg) monitors the country's most important economic sector with annual agricultural surveys covering just 2,000 households.
While these surveys have traditionally been used to estimate metrics such as yield and cropland, the high costs of ground data collection lead to a small sample size and limited insights. MinAg now aims to use these observations as training data, developing a SatML model to map agricultural variables at high-resolution nationally. However, these agricultural surveys are not designed for SatML, possibly limiting model performance. This reflects similar aims to use SatML to fill data gaps in low-income regions of the globe \citep{burke2021using}.
%
%
In Togo's case, 
the MinAg's next survey will include a small data augmentation budget aimed to optimize SatML performance. 
\emph{Where should those extra households be sampled, to maximize performance of the resulting model?} 
This question is particularly challenging as the cost of collecting additional samples varies with geography, reflecting high and uneven costs of accessing remote regions with limited public infrastructure.
%
\label{box:togo_setting}
\end{floatingexample}

\emph{(1) Available labeled data are not intended for downstream use in SatML training, hence they exhibit strong spatial biases or gaps}. Crowd-sourced or web-scraped data represent convenience samples, as observations tend to be more concentrated in populated or easily accessible areas. 
Even demographic and economic surveys carefully designed to be nationally or subnationally representative \citep{corsi2012demographic} often employ \emph{cluster sample} methodologies that result in sparse pockets of labeled data (\Cref{fig:data_setup_togo}).

\emph{(2) Budget constraints and travel costs limit new data collection.} In ground-referenced data collection, travel costs can be a significant factor, and some regions may be infeasible to sample altogether due to terrain or safety concerns.

Using this formulation, we \textbf{develop the first methods to optimize spatial training datasets for SatML prediction}, accounting for heterogeneous physical data collection costs and binding budget constraints. We are motivated in large part by a specific problem in Togo, where our co-authors need to know how to spend a modest budget to augment a planned survey for training a model to generalize across the entire country (Box 1).
In extensive simulation experiments across sampling settings in three continents, we find that our method outputs training datasets that can out-perform common sense baselines and status-quo sampling. This leads to either enhanced performance under a fixed sampling budget, or allows the user to keep performance standards high under increasingly stringent budget constraints.

With our problem formulation and novel method in hand, we can answer fundamental research questions on the nature of this problem, geared toward pressing questions that practitioners are already facing. 
Specifically, we design experiments to study: 
%
\begin{itemize}
\item What types of representation are important to training generalizable models with  spatial training data, and how do we optimize them in the face of budget constraints?
\item In what real-world  settings will additional data collection be most effective for increasing model performance?
\end{itemize}
We find that optimizing dataset design can have a significant impact on model performance, even when data collection budgets are tight. We hope this work provides practical tools for practitioners using SatML to fill critical data gaps and lays a foundation for future methodological work bridging the theory and practice of spatial dataset design in ML.



\subsection{Background \& Related work}

SatML models are increasingly used for high-stakes applications such as agricultural monitoring, disaster relief, and poverty mapping, yet rely on training labels that are unevenly distributed across space \citep{aiken2022machine, soman_can_2023, rolf2024mission}. Lack of representative, high-quality labels has been a consistent bottleneck in performance \citep{cai2022adaptive}: in a review of the use of satellite imagery and ML to fill gaps in economic surveys and agricultural and population censuses across the world, \citet{burke2021using}  emphasize that the ``largest constraint to model development is now training data rather than imagery." 

Prior approaches for collecting high-quality training data in SatML have cast this as an active learning problem, in which training datasets are refined through multiple rounds of label collection \citep{tuia_active_2009,Desai_2022_WACV,kellenberger2019half,lin2020active,rodriguez2021mapping,stumpf2013active,ghozatlou2024review}. In a highly related study (but not using satellite imagery), \citet{soman_can_2023} evaluate active learning methods for the task of poverty estimation from call data records and phone surveys in Togo. Active learning is well-suited for remote satellite image annotation or phone surveys, where labeling is fast, iterative, and cost-independent across images. In contrast, ground-referenced data collection requires careful pre-planning and is usually executed in a single round, making active learning methods less practical.

Our problem formulation for one-round dataset augmentation is closer to traditional optimal experiment design formulations \citep{pukelsheim2006optimal}, where data points (experiments) are chosen to satisfy desired optimality conditions. Related data valuation methods seek to measure the marginal utility of data points \citep{ghorbani_shapely_2019, wang_data_2023}, although these approaches typically focus on retroactively quantifying utility.  The approaches we develop are inspired by  recent work showing that dataset composition is a consistent determinant of ML model performance \citep{rolf_representation_2021,hashimoto21a}. These works indicate that obtaining training data from different demographic groups within a population or sources of training data can lead to large differences in accuracy. 

Here, we adapt these insights to spatial data, where groups can be defined in multiple ways -- and where costs of labeling data can vary by group and location -- to generate actionable conclusions regarding optimal sample design in SatML.

\section{Optimizing spatial training data composition under physical travel constraints}
Our goal is to maximize map accuracy\footnote{Our framework can easily be extended to other objectives, such as maximizing accuracy in certain sub-regions.} 
of SatML models over specified geographic extents by selecting high-quality training data under physical sampling constraints.
As there are many established SatML methods that can be used to associate collected labels with satellite imagery, we  focus our optimization on the spatial distribution of training data (\Cref{fig:data_setup_togo}, right).
Motivated by real-world settings, we model this as a \emph{data augmentation} problem: starting with some set of locations where labels are already sampled or will be sampled (e.g., a cluster sample from an annual survey), we aim to collect additional data to optimally extend that sample, under tight budget constraints.
To keep our problem formulation as general as possible, we assume that the augmented sample must be designed only with knowledge of the spatial distribution of the original sample -- this also reflects our motivating problem setting in Box 1, where the entire survey must be designed at once, before any labels are collected. 

\subsection{Problem formulation}

The SatML models we use are trained on image feature inputs $X = \{\mathbf{x}_1, \ldots, \mathbf{x}_n \}$ and labels $Y = \{ y_1, \ldots, y_n\}$, then used to make predictions across the target region, composed of prediction units $X_{\textrm{tgt}}$, which is typically a grid of satellite tiles or small administrative units covering an entire country. We differentiate between prediction units $X$ (satellite tiles), and \emph{sampling units}, denoted by $S = \{s_1, \ldots, s_m \}$, which represent the granularity at which data collection decisions are made. Sampling units can be any geographic units, for example census tracts, enumeration areas, or spatial clusters of households.  We allow partial sampling within units (e.g., some households within a village), as is common for surveys defined over clusters or administrative areas.
We can assume that each $\mathbf{x}_i \in X$ corresponds to only one $s$ and that the spatial correspondence between any $s$ and its $\mathbf{x}_i$ is known.

We denote the set of sampling units where data is already available or planned to be collected as $S_0$.
%
We will augment this sample by selecting an additional set \( S_L \) to be labeled, drawing from the ``source" set of possible sampling units \( S_{\mathrm{src}}\).
The total labeled training set is then all $\{\mathbf{x}_i,y_i\}$ pairs that correspond spatially with \(S_0 \cup S_L\).

Our goal is to pick $S_L$ to minimize the expected prediction loss of the trained SatML model averaged across the target region \( X_{\mathrm{tgt}} \), under a cost budget \( B \). The objective is:
\begin{equation}\label{eqn:objective}
\begin{gathered}
    \argmin_{S_L \subset S_{\mathrm{src}} \setminus S_0 } \ \ \mathbb{E} \left[ \sum_{\mathbf{x}_i \in X_{\mathrm{tgt}}} \ell(\hat{f}(\mathbf{x}_i), y_i) \right] \\
    \text{s.t.} \quad \hat{f} \sim M(S_0 \cup S_L), \\
    \text{and} \quad c(S_0 \cup S_L) \leq B
\end{gathered}
\end{equation}
where \( \ell : Y \times Y \to \mathbb{R}^+ \) is the loss function, $\hat{f}$ is trained using $M$ on $\{(\mathbf{x}_i, y_i) | \mathbf{x}_i \in S_0 \cup S_L\}$, and \( c : 2^S \to \mathbb{R}^+ \) assigns a cost to each subset of spatial units. In practice, sampling costs will depend greatly on the specific problem setting, so we leave this function in a general form. 

\Cref{eqn:objective} models the SatML predictor {\( \hat{f} : X \rightarrow Y \)} as an (unknown) stochastic function \( M \) of its training data. This
follows past work \citep{rolf_representation_2021}, where the authors estimate the expected loss of a trained predictor as a parametric function of (non-spatial) properties of the training data.
In practice, \cref{eqn:objective} is impossible to solve as $M$ is unknown. 
Thus, in order to optimize dataset design in practice, we also need to estimate the relationship between a training set and the expected loss of a model trained on that sample -- in deviation from past work, we propose to use utility functions as a proxy estimate of model performance.

\subsection{Method: Optimizing proxy quality measures}
Our proposed method optimizes a proxy utility function \( U \) defined over sets of sampling units:
\begin{equation}\label{eqn: opt}
\argmax_{S_L \subset S_{\mathrm{src}} \setminus S_0} \ U(S_0 \cup S_L) \quad \text{subject to} \quad c(S_0 \cup S_L) \leq B.
\end{equation}
A good utility function will serve as a proxy for training set quality, such that maximizers of \Cref{eqn: opt} also incur low values of the expected loss in \Cref{eqn:objective}.
Since we can design the utility function $U$ we use in our method, we select functions that make \Cref{eqn: opt} tractable to optimize.
%

We propose and test several concave utility functions, with the goal of understanding which proxy measures of quality best approximate the solution to \cref{eqn:objective} under heterogeneous budgets $B$ and cost functions $c(\cdot)$. 
Motivated by past work in data selection \citep{rolf_representation_2021}, we propose utility functions that capture two main factors of dataset composition: size and representativeness. We characterize representativeness through groups determined by geography, image embeddings, and other auxiliary information.  
We represent each of these as a function of a sample inclusion vector $\mathbf{s} \in \{0,1\}^{|S_{\textrm{src}}|}$, where $\mathbf{s}_i$ denotes whether sampling instance $s_i$ contributes to the labeled sample $S_0 \cup S_L$. When solving the optimization in practice, we allow continuous values $\mathbf{s} \in [0,1]^N$, which we treat as sampling probabilities.

Our first utility function reflects the common understanding that data quantity is important for model performance, and therefore is designed to maximize \textbf{dataset size}, where \( U(S) = |S| \). Optimizing for a dataset size objective results in greedy sampling of the least expensive spatial units first until the budget is reached.

For a sample $S$, let $n_g$ be the number of training instances belonging to each group $g$ in a finite set of groups $\mathcal{G}$. Let $\gamma_g$ denote the proportion of group $g$ in the population. 
    The (weighted) utility group-based representativeness utility function encourages balanced group coverage and is defined as 
    \[
        U(S) = -\lambda \sum_{g \in \mathcal{G}} \gamma_g  n_g^{-1/2} - (1 - \lambda) n^{-1/2} .
    \]
    The parameter $\lambda \in (0,1)$ balances group-based representativeness and overall dataset size, and follows from the group risk decomposition in \citet{rolf_representation_2021}. When $\lambda = 1$, this objective reduces to the same dataset size objective.   


To apply the concept of group-based representation to our spatial data optimization settings, we propose two broad strategies for defining groups. The first use existing administrative boundaries (e.g. states) as the set of groups $\mathcal{G}$. The second uses the distribution of images or other geographic data (e.g. land cover maps) to determine groups. 


While the size-based utility function is intuitive, group-based representation objectives additionally score candidate training sets based on how well they representat the region at large. In \Cref{fig:utility_vs_r2} (described in detail in the Experiments section), we find all of our proposed utility functions are relatively good measures for ranking training datasets by performance of the resulting models. 

\section{Experimental settings}
We use three datasets, each from different continents, representing agricultural and socioeconomic prediction tasks emblematic of SatML applications. All datasets have good spatial coverage over a country, allowing us to simulate sampling training datasets according to different methods. In our experiments, we randomly allocate 80\% of the available data as the source set from which we draw different training set samples, and use the remaining 20\% of the data as a test set to evaluate model performance across the country.

\subsubsection{Togo Soil Fertility} This dataset is closest to our motivating example and consists of 23,683 instances covering a 1km$^2$ grid of Togo, excluding protected areas, forests, and hydrographic zones. Each point is associated with a Sentinel-2 satellite image and labeled with water pH 
from a ground survey. Note that the data structure differs from the motivating example in Box 1, as here we require sufficiently dense labeled data to evaluate the performance of various sampling regimes (such evaluation is impossible with the standard, clustered, 2,000-household agricultural survey). The original dataset lacks temporal labels, so we trained our models using features derived from 10m resolution 10 band Sentinel-2 images across two temporal windows spanning six years. We used the best-performing features (using imagery from January to June 2021) for all our analyses.

\begin{table}[t]
\centering
\begin{tabular}{lcccc}

\toprule
\multirow{2}{*}{Sampling Type} & \multirow{2}{*}{Size} & \multicolumn{3}{c}{Rep (ours)}  \\
\cmidrule(lr){3-5}
 & & Admin & Image & NLCD \\
\midrule
Cluster & 0.740 & 0.740 & 0.744 & 0.736 \\
Convenience/Urban & 0.689 & 0.518 & 0.685 & 0.701 \\
Random & 0.959 & 0.716 & 0.955 & 0.949 \\
\midrule
Overall & 0.381 & 0.347 & 0.510 & 0.540 \\
\bottomrule
\end{tabular}
\caption{\textbf{Utility functions can estimate the performance of training datasets before labels are collected.} Spearman rank correlation ($\rho$) between utility metrics and $R^2$ scores of models trained on subsets of the USA Population dataset. $\rho$ values are provided both within sampling types, and aggregated across sampled sets of all types (``Overall").}
\label{tab:rho_values}
\end{table}

\begin{figure}[h]
\centering
  \begin{minipage}{0.55\columnwidth}
    \includegraphics[height=3.7cm]{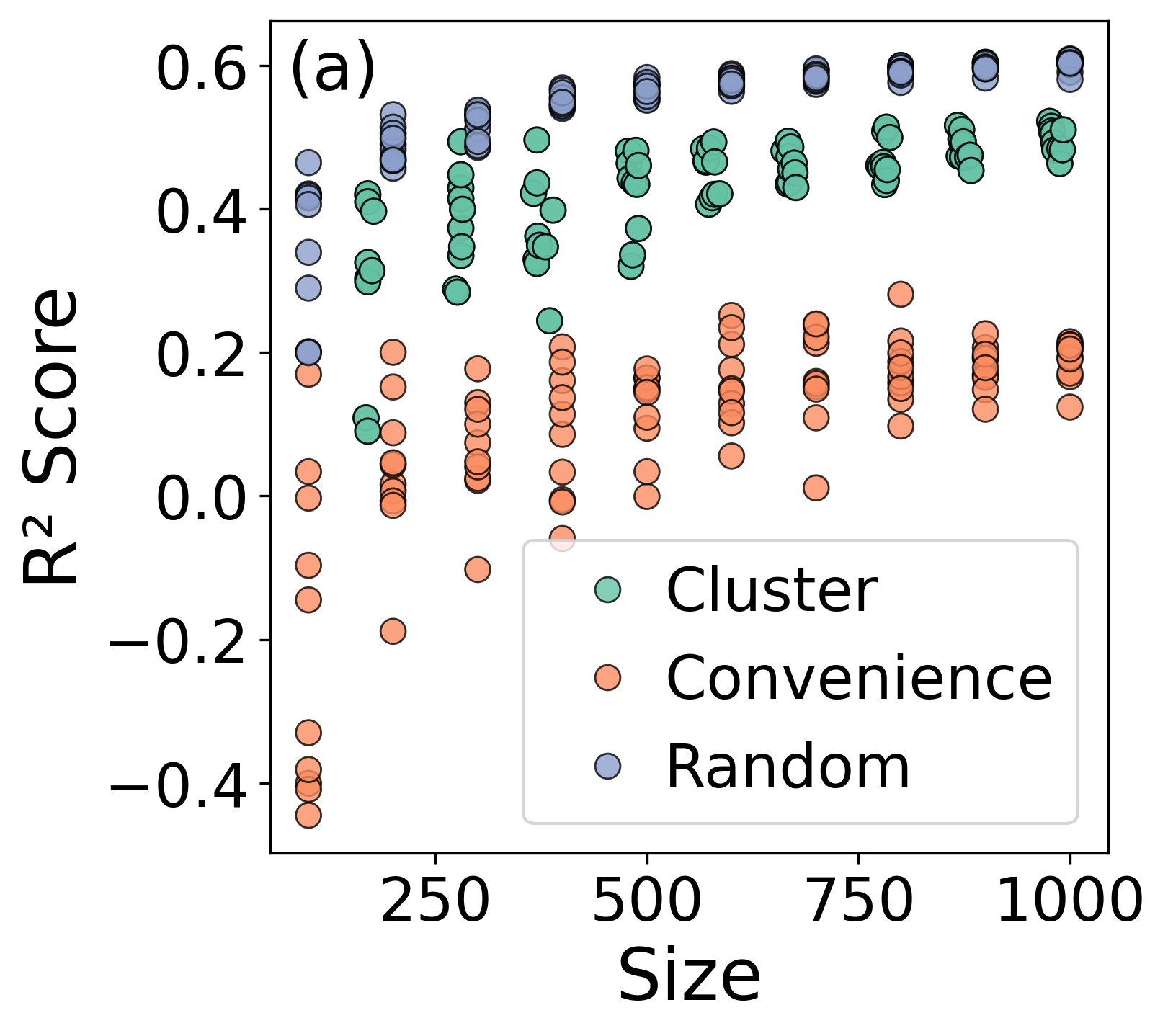}
  \end{minipage}
  \hfill
  \begin{minipage}{0.425\columnwidth}
    \includegraphics[height=3.7cm]{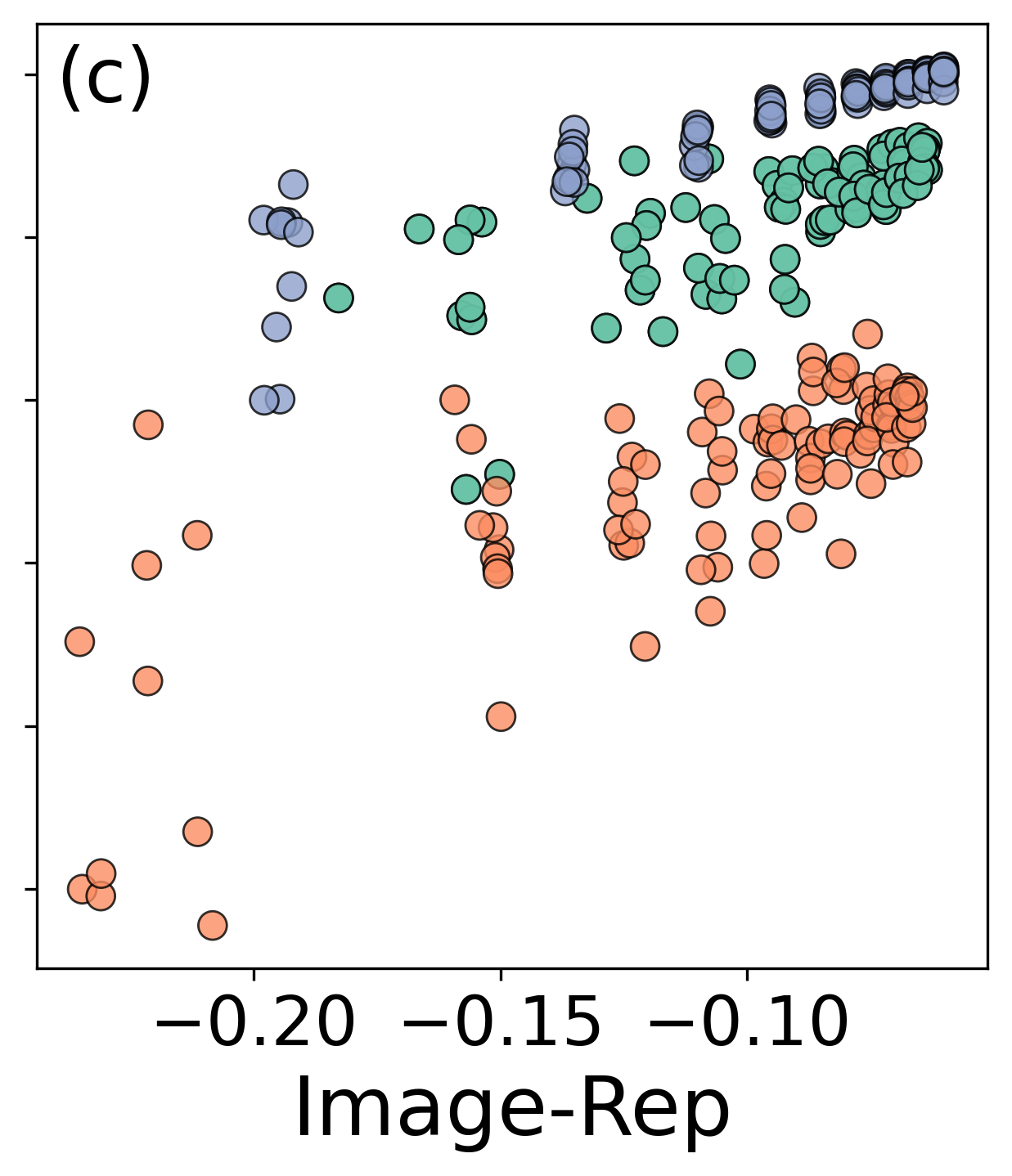}
  \end{minipage}
    \caption{ 
    \textbf{Illustrative examples of $R^2$ score vs. utility metric}, corresponding to dataset samples used in \Cref{tab:rho_values}.
    }
    \label{fig:utility_vs_r2}
\end{figure}

\subsubsection{India SECC Consumption} We use the per-capita consumption measurements from the 2012 Indian Socio Economic and Caste Census (SECC) to represent a developing country administrative data use case. The data were originally packaged as 
the Socioeconomic High-resolution Rural-Urban Geographic
Dataset for India (SHRUG) v2 (an updated version of \citet{asher2021development}) at the village/town level.
We use the dataset from \citet{aiken2023fairness}, which aggregates small administrative units together, resulting in 63,356 distinct units for training and evaluation. These units are matched to pre-computed MOSAIKS features derived from 4m resolution Planet Labs, Inc. imagery (accessed at \url{https://www.mosaiks.org/}
).

\subsubsection{USAVars Tree Cover,  Population Density} To obtain high-quality, densely-sampled data from a more data-rich environment, we use the USAVars dataset from \citet{rolf2021generalizable}. This dataset contains $97,876$ labeled 1 \textit{km}$^2$ grid cells, distributed roughly uniformly at random across the contiguous US. Labels are matched with features derived from NAIP aerial imagery (1 \textit{km}$^2$ crops of 1m resolution imagery resampled to 4m/pixel).
We use two different variables from USAVars -- percent tree cover and (log transformed) population density -- both of which are available on the torchgeo Python package \citep{torchgeo}. 
%

\subsubsection{MOSAIKS Image feature embeddings.} 
For all experiments, we use unsupervised random convolutional feature (RCF) embeddings with a trained ridge regression prediction head (MOSAIKS; \citet{rolf2021generalizable}) as our SatML method, fit using 5-fold cross-validation over the train set, with regularization parameter selected from 10 log-spaced values from $10^{-5}$ to $10^5$. 
MOSAIKS is designed to be computationally accessible, and thus represents the type of SatML model likely to be used in developing countries and resource-constrained settings. MOSAIKS has been shown to have high predictive skill \citep{brown2025alphaearthfoundationsembeddingfield,Corley_2024_CVPR}. This fits our use case of keeping the SatML model fixed while experimenting with different data conditions. Since MOSAIKS only requires retraining a linear model, we are able to experiment with many more data conditions and random seeds than with a fully trained neural network. 
For the India dataset, we use the same precomputed MOSAIKS features as  \citet{aiken2023fairness}; for the Togo and USAVars datasets, we generate new RCF features using the torchgeo \citep{torchgeo} RCF implementation with empirical patches of kernel size $4$ and bias $-1$.

\subsection{Conditions: initial samples and augmentation costs}\label{sec: prob settings}
%

Our experimental simulation protocol is designed to closely mimic existing survey protocols, particularly those conducted in developing countries, such as the Demographic and Health Surveys (DHS) Program. 
To draw the initial samples for each country, we sample clusters with probability proportional to their size from a subset of $N$ strata across the country; within each cluster we label $k$ points chosen uniformly at random. For clusters with fewer than $k$ points, we sample all points. 
To augment the initial set, we sample at the cluster level and label $k$ points uniformly at random. 
We fix the choice of $N$ strata but resample the clusters from within those $N$ strata each time, so our results reflect variability due to different possible initial samples after strata are determined. 
To reflect spatially heterogeneous costs, we assign cost $c_1$ to clusters that fall within the $N$ strata in the initial sample and $c_2$ to clusters in other strata, where $c_2 > c_1$  reflects higher costs of accessing unseen strata, due to required travel and training new enumerators. 


Due to the differing sizes of each country and the different structures of our three datasets, we tailor the parameters of our simulations to be meaningful for each country.
For the USA setting, the strata are the 48 contiguous U.S. states and clusters are counties within each state, $N=5$, and $k=10$.  
For the India setting, the strata are 28 states and clusters are districts within each state,  $N = 10$, and $k = 20$. 
%
For the Togo setting, the strata are Togo's 5 regions and the clusters are cantons within each region, $N = 2$, and  $k = 25$. 

\subsection{Group-based representation}

Across the three countries, we define two types of spatial groups: groups determined by administrative units, and image-based groups, determined by $k$-means clustering in high-dimensional image embedding space. We refer to corresponding utility functions as \textbf{Admin-Rep} and \textbf{Image-Rep}, respectively. In the USA and India settings, states serve as the administrative units in Admin-Rep, and images are clustered into 8 groups for Image-Rep. For Togo, we use regions as the administrative units and form 3 image groups. In the USA setting, we include an additional grouping using data from the National Land Cover Database (NLCD). At a resolution of 30m, each pixel in the 1km$^2$ NAIP image is assigned a land cover class. The resulting land cover proportion vectors are clustered using $k$-means in embedding space, resulting in $8$ NLCD groups; we refer to the corresponding utility function as \textbf{NLCD-Rep}. Both NLCD-Rep and Image-Rep construct groups that are clustered in data space, though not necessary spatially contiguous.

In maximizing the group-based representation, we experiment with $\lambda = 0.5$ and $\lambda = 1$, and report $\lambda = 0.5$. We solve this optimization using the MOSEK solver via cvxpy.

%


\begin{table*}[t!]
\centering
\renewcommand{\arraystretch}{1.1} 
\begin{tabular}{llcccccc}%
\toprule
  & & \multicolumn{6}{c}{method used to augment initial sample} 
\\ \cmidrule(lr){3-8}
\multicolumn{1}{l}{ \shortstack[l]{\textbf{Country \& Task} \\ (Initial sample $R^2$) }} & Budget 
& \multicolumn{1}{c}
{Default}
& \multicolumn{1}{c}
{Greedy (Size)}
& \multicolumn{1}{c}
{Random}
& \multicolumn{1}{c}{\shortstack{Rep. (ours) \\ admin}}
& \multicolumn{1}{c}{\shortstack{Rep. (ours) \\ image}}
& \multicolumn{1}{c}{\shortstack{Rep. (ours) \\ NLCD}} \\
\hline
\multirow{4}{*}{\shortstack[l]{\textbf{Togo Soil Fertility} \\($0.09 \pm 0.12$)}}  & 100 & 0.10 ± 0.13 & 0.08 ± 0.14 & 0.13 ± 0.07 & 0.08 ± 0.15 & \textbf{0.17 ± 0.04} & n/a\\%
& 200 &  0.15 ± 0.05 & 0.06 ± 0.17 & 0.12 ± 0.11 & 0.15 ± 0.07 & \textbf{0.18 ± 0.04} & n/a\\%
& 500 & 0.15 ± 0.06 & 0.06 ± 0.11 & 0.17 ± 0.03 & 0.20 ± 0.04 & \textbf{0.21 ± 0.02} & n/a\\%
& 1000 & 0.15 ± 0.07 & 0.13 ± 0.05 & 0.20 ± 0.05 & \textbf{0.23 ± 0.02} & \textbf{0.23 ± 0.01} & n/a\\%
\hline%
\multirow{4}{*}{\shortstack[l]{\textbf{India Consumption} \\ ($0.13 \pm 0.06$)}} & 500 & 0.14 ± 0.04 & 0.14 ± 0.06 & \textbf{0.18 ± 0.09} & \textbf{0.18 ± 0.08} & 0.17 ± 0.04 & n/a\\%
& 1000 & 0.16 ± 0.05 & 0.18 ± 0.04 & \textbf{0.26 ± 0.05} & 0.19 ± 0.27 & 0.18 ± 0.06 & n/a\\%
& 2000 & 0.18 ± 0.04 & 0.16 ± 0.10 & 0.29 ± 0.06 & \textbf{0.33 ± 0.02} & 0.28 ± 0.02 & n/a\\%
& 5000 & \emph{infeasible} & 0.31 ± 0.07 & 0.37 ± 0.02 & 0.38 ± 0.03 & \textbf{0.39 ± 0.02} & n/a\\%
\hline%
\multirow{4}{*}{\shortstack[l]{\textbf{USA Population} \\  ($0.0 \pm 0.26$)}} & 50 & \textbf{0.25 ± 0.19} & -0.21 ± 0.60 & 0.20 ± 0.16 & 0.10 ± 0.34 & 0.10 ± 0.30 & 0.13 ± 0.26\\%
& 100 & 0.31 ± 0.11 & 0.25 ± 0.17 & 0.23 ± 0.12 & 0.25 ± 0.12 & \textbf{0.35 ± 0.07} & 0.28 ± 0.11\\%
& 500 & 0.45 ± 0.03 & 0.44 ± 0.04 & 0.45 ± 0.05 & \textbf{0.47 ± 0.03} & 0.45 ± 0.03 & 0.46 ± 0.01\\%
& 1000 & \emph{infeasible} & 0.48 ± 0.04 & 0.50 ± 0.03 & \textbf{0.52 ± 0.02} & 0.49 ± 0.04 & \textbf{0.52 ± 0.01}\\%
\hline%
\multirow{4}{*}{\shortstack[l]{\textbf{USA Treecover}  \\ ($0.49 \pm 0.13$)}} & 50 & 0.56 ± 0.10 & 0.62 ± 0.10 & 0.60 ± 0.10 & 0.55 ± 0.12 & 0.57 ± 0.06 & \textbf{0.68 ± 0.05}\\%
& 100 & 0.63 ± 0.07 & 0.63 ± 0.07 & 0.65 ± 0.08 & \textbf{0.68 ± 0.05} & 0.66 ± 0.06 & 0.66 ± 0.09\\%
& 500 & 0.75 ± 0.04 & 0.75 ± 0.03 & 0.72 ± 0.05 & 0.77 ± 0.01 & 0.76 ± 0.02 & \textbf{0.78 ± 0.02}\\%
& 1000 & \emph{infeasible} & 0.79 ± 0.01 & 0.78 ± 0.04 & \textbf{0.81 ± 0.01} & \textbf{0.81 ± 0.01} & \textbf{0.81 ± 0.01}\\%
\bottomrule
\end{tabular}%
\caption{Prediction performance ($R^2$ scores $\pm$ one standard deviation for 10 random seeds), of SatML models trained with augmented samples. 
%
Here, budget represents the additional budget due only to augmentation. Initial samples of size 500 are chosen for Togo, size 2000 are chosen for India consumption, and size 100 are chosen for USA tasks. 
Since NLCD data is only available in the U.S., for Rep-NLCD is not applicable (n/a) for Togo and India. The default clustering performance is excluded for large budgets in India and the USA as these budgets cannot be realized with the cluster sampling procedure .}
\label{tab: augmentation}
\end{table*}

\section{Experiments}

\subsection{Estimating the quality of spatial 
training data}
Before we optimize spatial data collection in the face of physical data collection costs, we first conduct an experiment to isolate the capabilities of our utility functions. As we want to use these functions for optimization objectives, we are specifically interested in how well each utility function can rank different spatial training datasets in terms of the performance of a resulting trained SatML model.  

In \Cref{tab:rho_values} and \Cref{fig:utility_vs_r2}, we analyze the $R^2$ performance of a SatML model trained on different datasets in relation to the utility of each dataset. The setup is the same as in \Cref{fig: data collection initial set performance}: we take sub-samples of the USAVars Population training set, at increments of 100 points from 100 to 1000 using three sampling methods: cluster sampling  with 10 points per cluster, convenience sampling, in which samples are selected with probability proportional to distance from the top 20 urban areas in the USA, and random sampling over space.

All of our proposed utility functions demonstrate skill at ranking training different datasets in terms of the downstream performance, as indicated by positive Spearman $\rho$ rank correlation coefficient values. Among these, the group-based utility functions Image-Rep and NLCD-Rep most closely align (in rank) to $R^2$ performance. All utility functions capture meaningful rankings within each specific sample type (Random, Cluster, Convenience), as indicated by similar Spearman's $\rho$ values (with $\rho > 0.5$). However, there is more variability in predictive value when such utility functions are used to rank all the training sets in aggregate, with Image and NLCD-based representation outperforming size and administrative boundary representation metrics.


\subsection{Evaluating our methods for spatial data collection}

\begin{figure}[thb]
    \centering
\includegraphics[width=.9\linewidth]{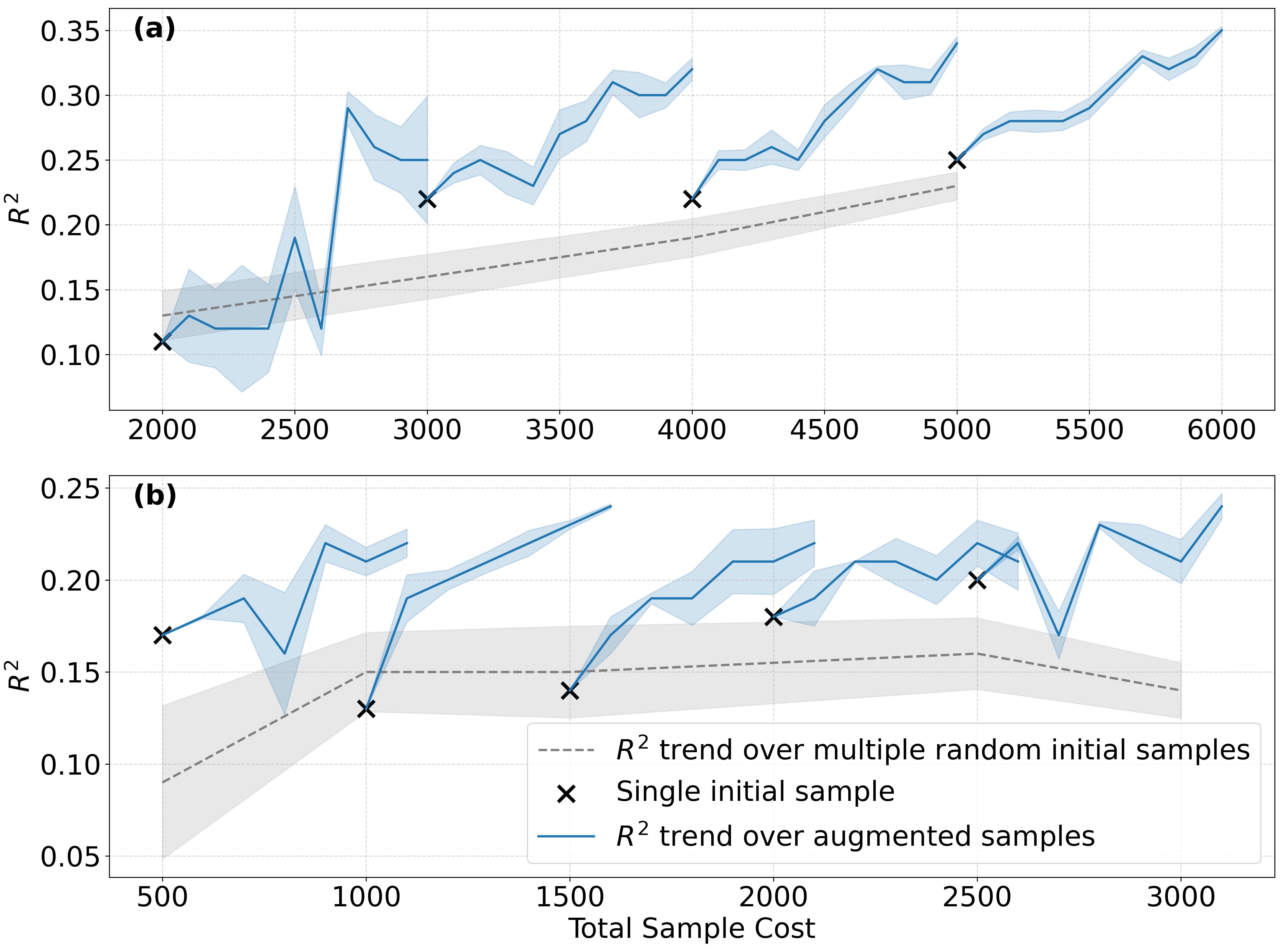}
\caption{\textbf{Optimized sampling improves performance compared to default cluster sampling methods.} Performance of our best-performing cost-aware augmentation methods for India (Rep-Admin, panel a) and Togo (Rep-Image; panel b). Blue lines show augmentations to a a single initial sample (black $\times$'s). Gray dashed line and shaded areas show mean $\pm$ 1 standard error across 10 random  initial samples. Augmentation methods are run for 5 random seeds. 
     }
     \label{fig: varying_initial_set_size}
\end{figure}

Next, we assess progress toward our main goal --   optimizing these utility functions to increase model performance while adhering to spatial cost constraints. 
%
%
To evaluate our framework in these augmentation experiments, we use the cluster sampling method discussed in the previous section to generate initial samples of a fixed size that yield a positive (sometimes small) $R^2$ score. We construct these initial samples to be large enough to accurately compute standard metrics used in policy or reporting, though not necessarily sufficient for high model performance. We then simulate an additional round of data collection using small budgets.

\Cref{tab: augmentation} shows the resulting model performance when using Admin-Rep, Image-Rep, and NLCD-Rep, compared against 3 baselines: default cluster sampling, which samples clusters in the already-sampled strata with probability proportional to size, greedy clusters\footnote{We implement Greedy Clusters using  the size-based utility function to be consistent with the other optimization methods, yet still reflect the common heuristic of favoring large training datasets.}, which maximizes the number of clusters collected, and random clusters, which randomly selects clusters available for labeling regardless of cost.
We observe that our cost-aware optimized sampling methods present competitive strategies compared to baseline methods -- achieving the best average performance in nearly all settings. Overall, the optimized sampling framework is able to consistently improve over the baseline methods, with all three of our representation-based utility functions performing well compared to baselines. 
Notably, Image-Rep consistently achieves the highest performance on the Togo Soil Fertility dataset, though the winning representation-based method is more mixed in the other three tasks. 

While performance differences due to different sampling methods are often within standard deviations of each other, it is encouraging that our methods perform consistently well and outperform baseline methods in almost all datasets and budgets. 
In fact, the variability in performance indicated by the large standard deviations underscores the sensitivity of model performance to sampling methods and thus the importance of studying dataset design.
%

Interestingly, Random Clusters emerges as a competitive baseline in \Cref{tab: augmentation}. Particularly, for the India Consumption and Togo soil fertility tasks, Random Clusters tends to outperform Greedy Clusters and default Clusters, suggesting that there is fundamental value in using a sample that is well-dispersed over space, even at the cost of fewer samples.

\subsection{When is optimized data collection most valuable?} 

\begin{figure}[h]
    \centering
    \includegraphics[width=1.0\linewidth]{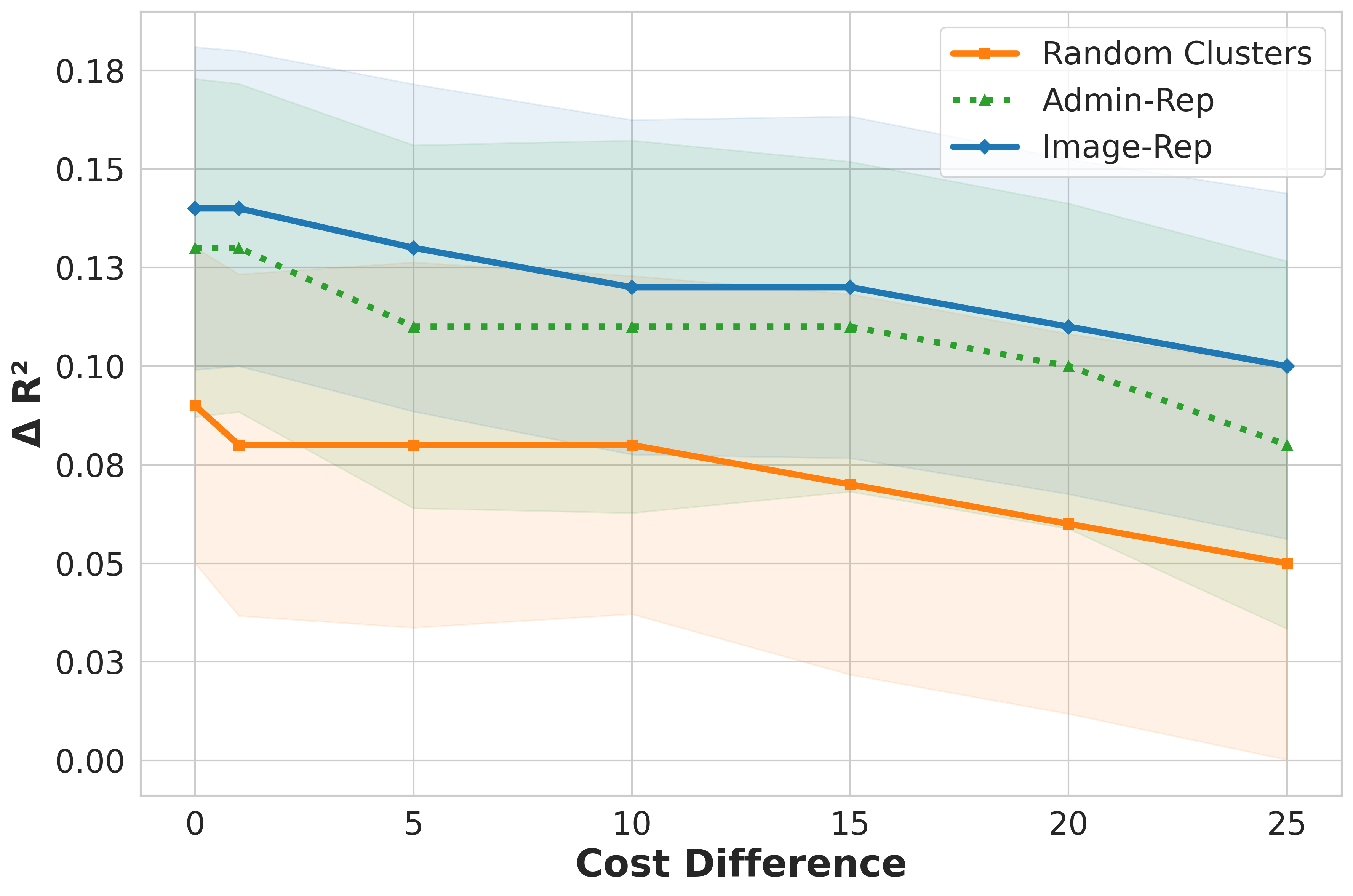}
    \caption{\textbf{Performance gains due to optimized sampling depend on the cost structure.} Performance gain ($\Delta R^2$ from initial sample) as we change the difference between in-region and out-of-region costs ($c_2 - c_1$) for the soil fertility prediction task in Togo. We use an initial samples of size 500, in-region cost of $c_1=25$, and a budget of 500. We increase the out-of-region cost $c_2$ between 25 and 50. Lines and shaded areas show average values $\pm$ 1 standard error.
    }
    \label{fig:varying_cost}
\end{figure}

Our next set of experiments investigate for what spatial problem settings is optimized sampling for augmentation worthwhile -- expanding our analysis from the tables above to assess performance across varying cost structures (\Cref{fig:varying_cost}), and different sizes of initial datasets (\Cref{fig: varying_initial_set_size}). We focus this analysis on our two datasets reflecting more real-world scenarios, India consumption and Togo soil fertility.

In \Cref{fig: varying_initial_set_size},  we compare optimized sampling to a status quo baseline, representing the possibility of governments or firms increasing sampling budgets, but allocating those budgets according to the same cluster strategy originally planned. Here, we model this as using larger budgets to allow more clusters to be sampled. We note that for large additional budgets, survey designers might add additional strata (increase $N$), but in these experiments, we focus on low-budget augmentation. Across both the datasets in \Cref{fig: varying_initial_set_size}, using our cost-aware augmentation methods (blue) results in performance gains over increasing the budget of default cluster sampling (grey) regardless of the initial sample size. Default cluster sampling is spatially constrained to predetermined strata, limiting its maximum performance. The augmented samples in both datasets consistently achieve a higher performance ceiling, even at lower total cost.


%

In addition to varying the initial samples, we investigate the sensitivity of our methods to differences in spatial cost structures. In \Cref{fig:varying_cost}, we vary the cost difference between in- and out-of-region cost for the Togo soil dataset. 
We find that performance gains ($\Delta R^2$) decrease as the cost of difficult-to-sample regions increases. This makes sense as higher cost regions result in less data being collected for the same budget. Even so, Image-Rep (blue) consistently offers higher average performance gains irrespective of the cost difference, demonstrating that cost-aware optimized sampling is robust to varying cost discrepancies across samples.



\begin{floatingexampletwo}[float, floatplacement=b]{Takeaways for the Togo problem setting}
Our results have several implications that can inform how to best augment the next year's planned agricultural surveys in Togo addressing our motivating problem setting outlined in Box 1. 
Overall, \textbf{our findings imply an image-based representation sampling method is recommended}, as the samples generated according to this utility function consistently performed best in Togo (\Cref{tab: augmentation}).
%
While default cluster sampling or choosing clusters randomly is within the margin of error bounds of our best results in \Cref{tab: augmentation}, we find that image-based sampling consistently performs well across different initial sampling budgets and data collection cost ratios (\Cref{fig:varying_cost,fig: varying_initial_set_size}), further increasing our confidence in recommending this method.
While our results are promising, we recommend proceeding with care when putting this method into practice for real-world, high-stakes data collection. 
First, we recommend visualizing the augmented samples and discussing their properties in consultation with the MinAg. 
Second, the annual survey collects data about different tasks than those used for our simulation study (due to data availability); thus, we recommend conducting preliminary analyses on related datasets if possible, to explore how the performance improvements recovered here are likely to translate.
\label{box:togo_takeaways}
\end{floatingexampletwo}

\section{Discussion}
The need for methods to aid in ground-based data collection for SatML is clear -- 
to quote \citet{burke2021using}, ``expanding the quantity, and in particular, the quality of labels will quickly accelerate progress in this field [of using satellite data for sustainable development] and will allow both researchers and practitioners to measure new outcomes and to accurately assess model performance.''
In this work, we present the first approach to optimized sampling in realistic settings of  preexisting clustered data and heterogeneous label collection costs. 
In experiments covering Togo, India, and the US, we found that our sampling optimization methods that encourage representation across spatial groups are almost always the top performing sampling method. 
The strong performance of our method, as well as the competitiveness of the baseline random clusters method, emphasizes the value of a representative or spatially dispersed sample, a paradigm that can be in conflict with the traditional goal in ML of collecting a large quantity of data. 


By establishing the important influence of spatial training data distributions on SatML model performance, our study has important implications for future work, including the immediate use of our results in Togo (discussed in Box 2) and promising areas for new research. 
Perhaps most prominently, there is the opportunity to improve on the optimization methods proposed here. Our results showed that the best way to define spatial groups may differ per country and prediction task; future work could work to provide a single best definition or delineate when different utility functions would be best. Testing data augmentation methods in settings where the initial sample has more extreme spatial biases (e.g. convenience sampling near population dense areas, which \Cref{fig: data collection initial set performance} suggests is bad for SatML) would also be an interesting and valuable area of future work.

\section{Acknowledgements}

We would like to thank the Togolese Ministry of Agriculture, Village Hydraulics and Rural Development as well as the Togolese Institute of Agronomic Research, for providing us with the Togo Soil Fertility Data; Sean Luna McAdams for facilitating collaboration; and Emily Aiken for assisting with access to the India SECC Data. This work was funded by The McGovern Foundation and the Fund for Innovation in Development. In this work, we also used Jetstream2 at Indiana University through allocation CIS240692 from the Advanced Cyberinfrastructure Coordination Ecosystem: Services \& Support (ACCESS) program, which is supported by National Science Foundation grants \#2138259, \#2138286, \#2138307, \#2137603, and \#2138296.

\bibliography{main}

\clearpage
\onecolumn
\appendix

\section{Appendix: Data}

\subsection{Togo Soil Fertility Data}

\begin{figure}[htbp]
    \centering
    \begin{subfigure}[b]{0.2\columnwidth}
        \includegraphics[width=\linewidth]{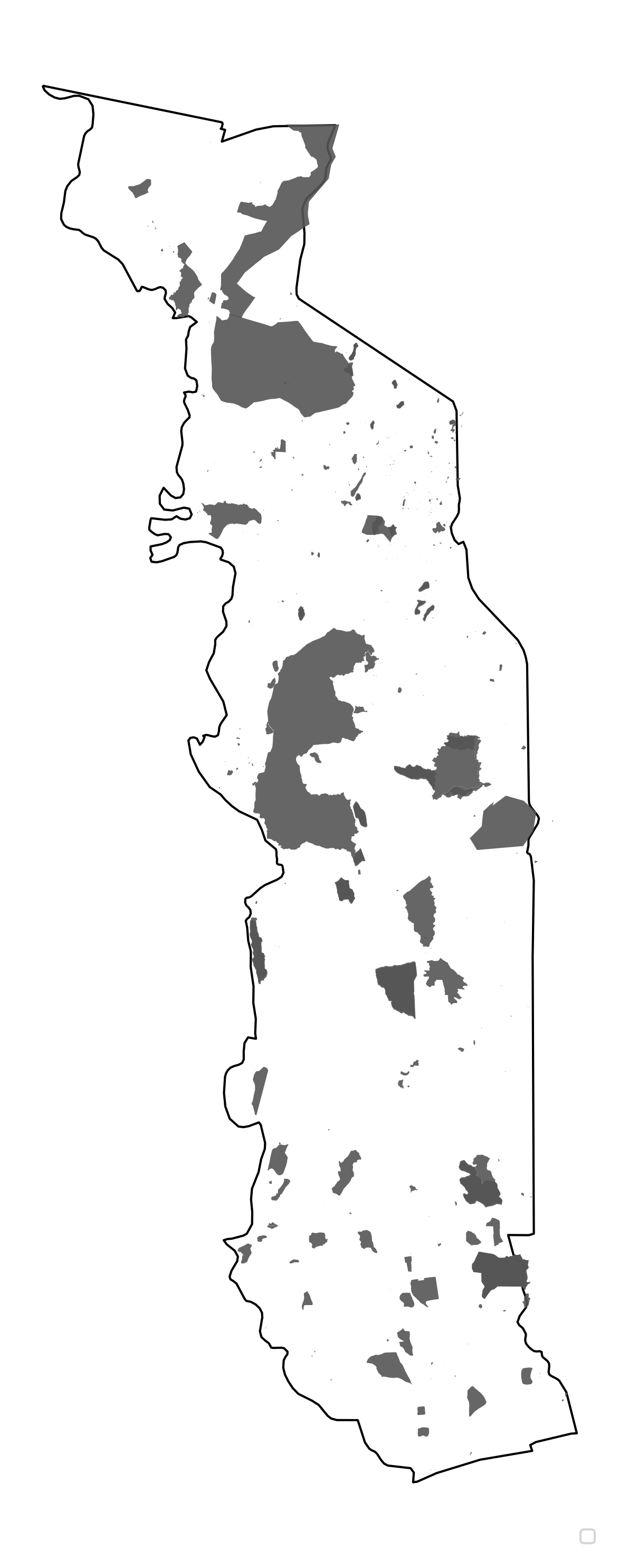}
        \caption{Protected Areas in Togo}
        \label{fig:protected}
    \end{subfigure}
    ~
    \begin{subfigure}[b]{0.2\columnwidth}
        \includegraphics[width=\linewidth]{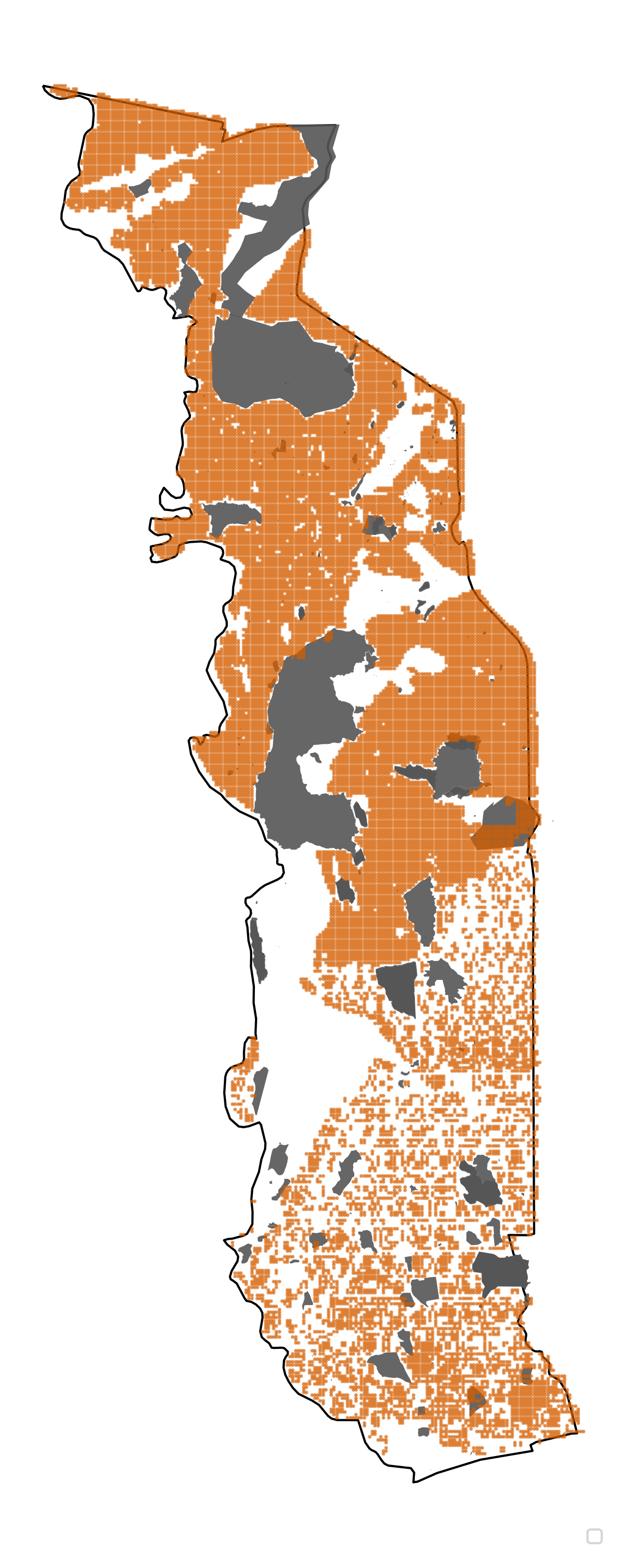}
        \caption{Soil Fertility Samples}
        \label{fig:soil}
    \end{subfigure}
    \caption{Maps of protected regions and sampled geolocations of soil fertility data across Togo}
    \label{fig:togo_maps}
\end{figure}

The Togo Soil Fertility Data consists of $23,683$ points across Togo of a 10 band Sentinel-2 image corresponding with a value for water pH. These points lie on a 1 km$^2$ grid covering the country except for protected areas and a mountainous region in the southwest of the country (see \Cref{fig:togo_maps}).

\section{Appendix: Experimental Details}

\subsection{MOSAIKS Image feature embeddings}

The India dataset, sourced from \citet{aiken2023fairness}, uses 4000-dimensional MOSAIKS image feature embeddings. For the USAVars and Togo datasets, we generate features using the torchgeo RCF implementation, generating 4096-dimensional features for USAVars and 4000-dimensional features for Togo.

\subsection{Sampling according to probabilistic inclusion vector}
Solving the optimization problem results in a vector $\mathbf{s} \in [0,1]^N$, where each element represents the sampling probability of the corresponding sample. To use these sampling probabilities, we randomly shuffle these indices, then perform binomial trials according to the probability to decide which samples to include, accounting for the cost of including the resulting samples. The process stops once the cumulative cost exceeds the budget, and the sampled points are used.

\subsection{Generating convenience samples}
To generate the convenience samples in \Cref{fig: data collection initial set performance} and \Cref{fig:utility_vs_r2}, we used the top 20 urban areas as defined by the 2020 U.S. Urban Areas Census \citep{census_urbanrural}. We computed the distances from each point to the nearest urban area and then applied max-min normalization. Then, we applied a soft-max to these values with temperature parameters $\tau=0.025$ for Population task and $\tau=0.001$ for the Tree Cover task. Since the Tree Cover task achieves a higher overall $R^2$ score, by reducing the temperature, we're able to simulate this extreme convenience sampling scenario as we did in the Population task.

\section{Appendix: Additional Results}

\subsection{Estimating the quality of spatial training data}

\begin{figure*}[t]
\centering
\includegraphics[height=4cm]{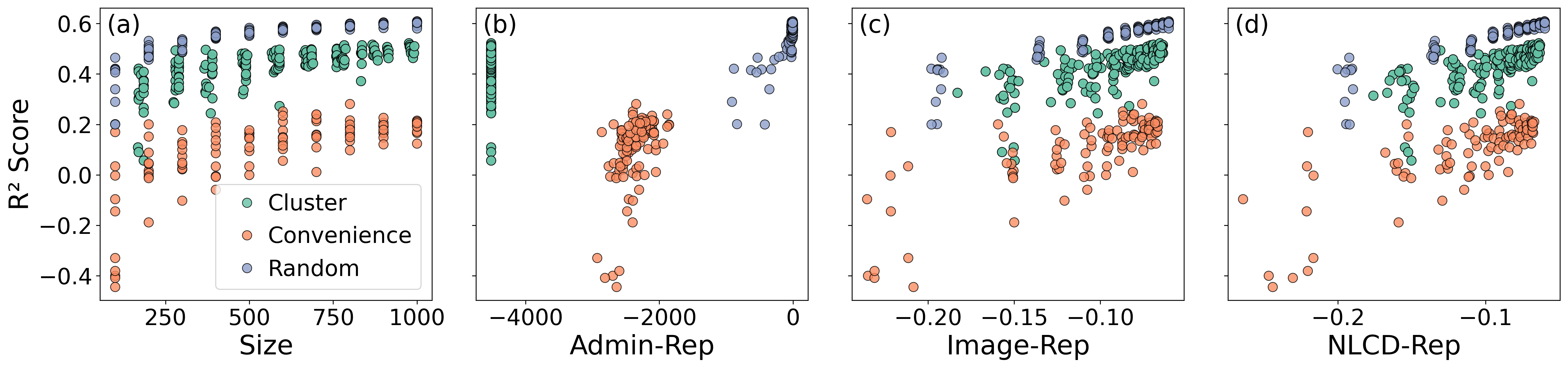} 

\vspace{0.3cm}

\includegraphics[height=4cm]{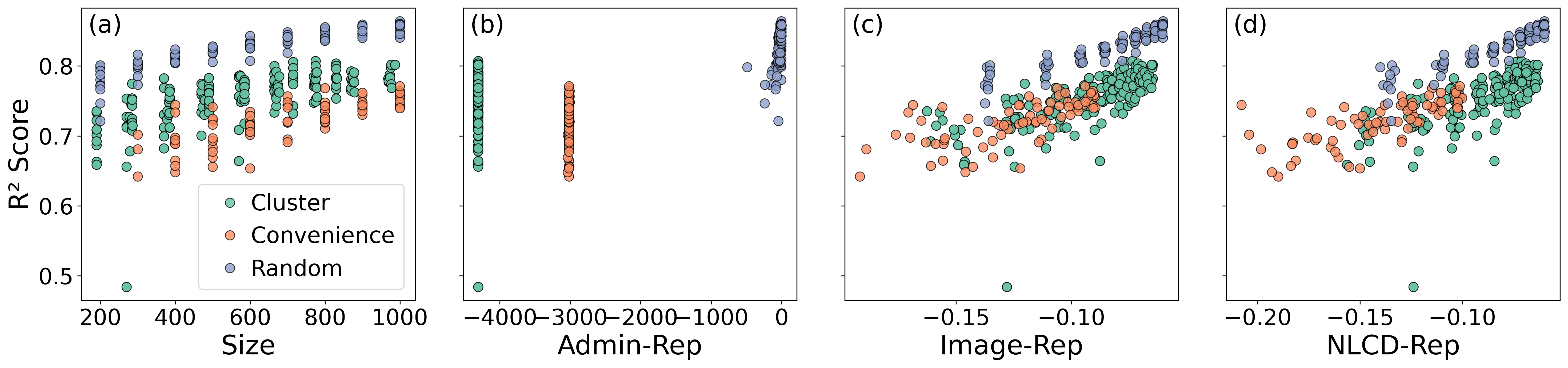}
\caption{
\textbf{Utility functions can estimate the performance of training datasets before labels are collected.}  
Top: USAVars Population task. Bottom: Tree Cover task. Each scatterplot shows the relationship between utility (x-axis) and model performance (\(R^2\), y-axis). 
For the Population task and Tree Cover task, convenience samples are generated with different parameters (See \textit{Generating convenience samples}). For the Population task, all data points are shown. For the Tree Cover task, only the top 95\% of utility values is plotted to improve visibility due to outliers with low utility.}

\label{fig:utility_vs_r2_pop_and_tree_all}
\end{figure*}

\Cref{fig:utility_vs_r2_pop_and_tree_all} and \Cref{tab:rho_values_treecover} provide additional results for the estimating the quality of spatial data with utility functions by including additional plots for the USAVars Population task (\Cref{fig:utility_vs_r2_pop_and_tree_all}), as well as plots and $\rho$ values for the Tree Cover task (\Cref{fig:utility_vs_r2_pop_and_tree_all} and \Cref{tab:rho_values_treecover}).

We only use the USAVars dataset to evaluate the effectiveness of these utility functions in ranking dataset quality, simulating the realistic scenario in which existing data is leveraged to select appropriate utility functions, prior to deploying the sampling framework on real-world datasets represented by the India and Togo data.

\begin{table}[b]
\centering
\begin{tabular}{lcccc}

\toprule
\multirow{2}{*}{Sampling Type} & \multirow{2}{*}{Size} & \multicolumn{3}{c}{Rep (ours)}  \\
\cmidrule(lr){3-5}
 & & Admin & Image & NLCD \\
\midrule
Cluster & 0.761 & 0.762 & 0.771 & 0.759 \\
Convenience (Urban) & 0.870 & 0.828 & 0.843 & 0.872 \\
Random & 0.959 & 0.773 & 0.963 & 0.953 \\
\midrule
Overall & 0.587 & 0.454 & 0.775 & 0.767 \\
\bottomrule
\end{tabular}
\caption{
\textbf{Utility functions can estimate the performance of training datasets before labels are collected.} 
Spearman rank correlation ($\rho$) between utility metrics and \(R^2\) scores of models trained on subsets of the USA Tree Cover dataset. 
Unlike the scatterplots in \Cref{fig:utility_vs_r2_pop_and_tree_all}, these correlations are computed over the full set of initial samples, including outliers.
}
\label{tab:rho_values_treecover}
\end{table}

\end{document}